\def\year{2022}\relax
\documentclass[letterpaper]{article} 
\usepackage{aaai22}  
\usepackage{times}  
\usepackage{helvet}  
\usepackage{courier}  
\usepackage[hyphens]{url}  
\usepackage{graphicx} 
\usepackage{natbib}  
\usepackage{caption} 
\DeclareCaptionStyle{ruled}{labelfont=normalfont,labelsep=colon,strut=off} 
\usepackage{algorithm}
\usepackage{algorithmic}

\usepackage{newfloat}
\usepackage{listings}
\floatstyle{ruled}
\newfloat{listing}{tb}{lst}{}
\floatname{listing}{Listing}

\usepackage{graphicx}
\usepackage{amsmath,amssymb}

\usepackage{color}
\usepackage{subfigure}

\usepackage{array}
\usepackage{mathtools}
\usepackage{multirow}
\usepackage[english]{babel}
\usepackage{nopageno}
\usepackage{diagbox}

\title{Elastic-Link for Binarized Neural Network}
\author {
    Jie Hu\textsuperscript{\rm 1},
    Ziheng Wu\textsuperscript{\rm 3},
    Vince Tan\textsuperscript{\rm 4},
    Zhilin Lu\textsuperscript{\rm 2},
    Mengze Zeng\textsuperscript{\rm 4},
    Enhua Wu\thanks{corresponding author}\textsuperscript{\rm 1,5}
}
\affiliations {
    \textsuperscript{\rm 1} State Key Lab of Computer Science, ISCAS \& University of Chinese Academy of Sciences\\
    \textsuperscript{\rm 2} Department of Electronic Engineering, Tsinghua University\\
    \textsuperscript{\rm 3} Alibaba Group\\
    \textsuperscript{\rm 4} ByteDance Inc.\\
    \textsuperscript{\rm 5} University of Macau\\ 
    hujie@ios.ac.cn, zhoulou.wzh@alibaba-inc.com, vince.tan@bytedance.com, zeitzmz@gmail.com, ehwu@um.edu.mo
}

\begin{document}

\maketitle

\begin{abstract}

Recent work has shown that Binarized Neural Networks (BNNs) are able to greatly reduce computational costs and memory footprints, facilitating model deployment on resource-constrained devices. However, in comparison to their full-precision counterparts, BNNs suffer from severe accuracy degradation. Research aiming to reduce this accuracy gap has thus far largely focused on specific network architectures with few or no $1\times1$ convolutional layers, for which standard binarization methods do not work well. Because $1\times1$ convolutions are common in the design of modern architectures (e.g. GoogleNet, ResNet, DenseNet), it is crucial to develop a method to binarize them effectively for BNNs to be more widely adopted. In this work, we propose an ``Elastic-Link'' (EL) module to enrich information flow within a BNN by adaptively adding real-valued input features to the subsequent convolutional output features. The proposed EL module is easily implemented and can be used in conjunction with other methods for BNNs. We demonstrate that adding EL to BNNs produces a significant improvement on the challenging large-scale ImageNet dataset. For example, we raise the \mbox{top-1} accuracy of binarized ResNet26 from 57.9\% to 64.0\%. EL also aids convergence in the training of binarized MobileNet, for which a top-1 accuracy of 56.4\% is achieved. Finally, with the integration of ReActNet, it yields a new state-of-the-art result of 71.9\% top-1 accuracy.
\end{abstract}

\section{Introduction}

Convolutional Neural Networks (CNNs) have led to a series of breakthroughs for a variety of visual tasks~\cite{krizhevsky2012alexnet, long2014segmentation, ren2015frcnn, toshev2014humanpose, zhu2016action}. However, the challenge of resource constraints in terms of latency and memory storage is often faced when deploying CNNs on mobile or embedded devices. Previous work~\cite{cai2017hwgq, jacob2018integer, mckinstry2018faq, sambhav2019tqt, wu2016wage, sung2015l2, yang2019quantization, zhou2016dorefa} has demonstrated that quantizing the real-valued weights and activations of CNNs into low-precision representations can reduce memory footprint while still achieving good performances. This class of methods allows fixed-point arithmetic to be applied, which substantially accelerates inference and reduces energy costs. Taken to an extreme, both the weights and the activations can be represented with binary tensors \{-1, +1\}. Such networks are termed Binarized Neural Networks (BNNs)~\cite{courbariaux2016bnn}. In BNNs, arithmetic operations for convolutions can be replaced by the more efficient \textit{xnor} and \textit{bitcount} operations.

\begin{figure}
\centering
\includegraphics[width=0.35\textwidth,trim=80 125 70 70]{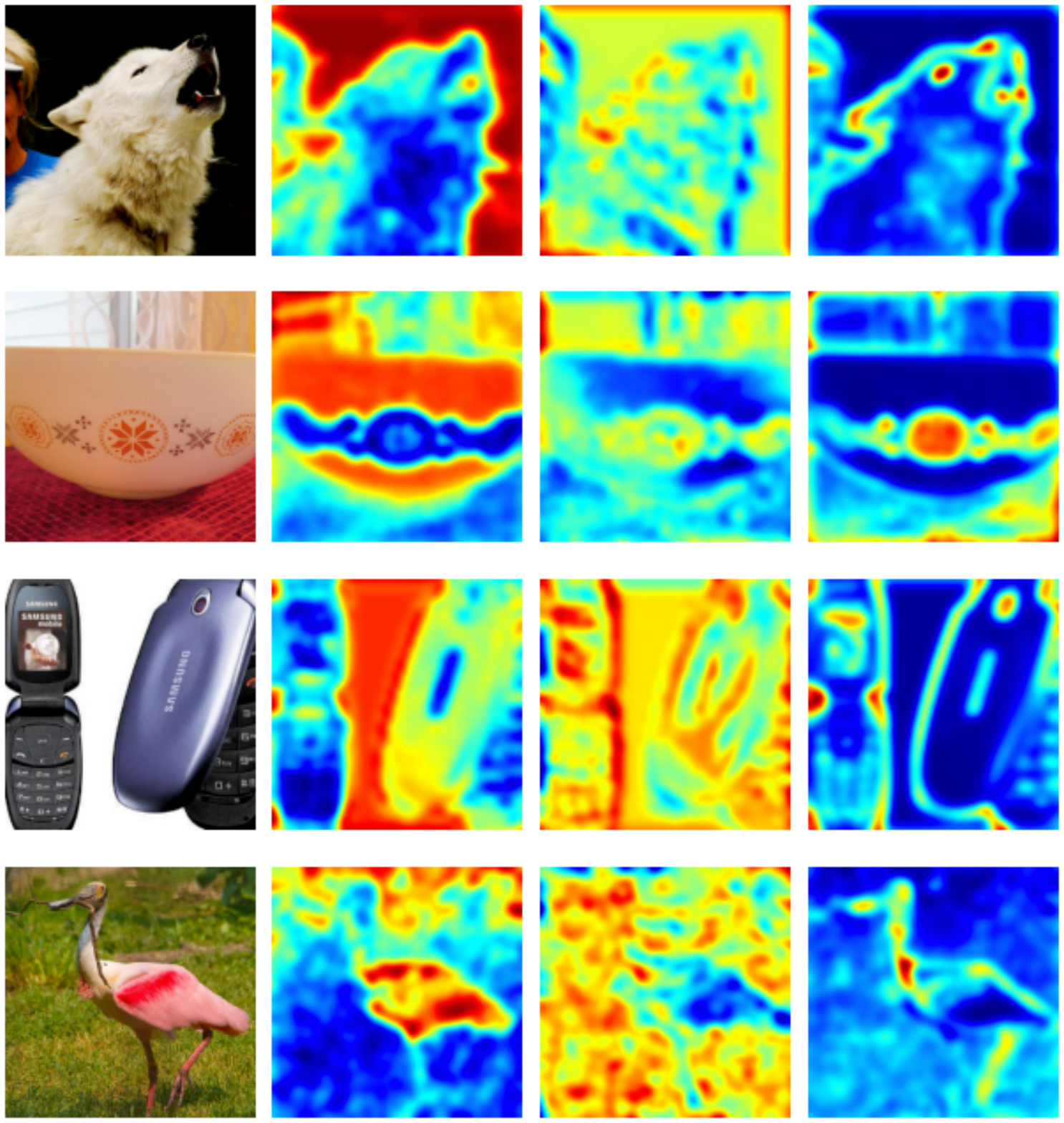}
\caption{Example images illustrating the same features on full-precision ResNet26 (2nd column), Bi-Real-ResNet26 (3rd column) and our proposed EL-ResNet26 (4th column).}
\label{fig:heatmap}
\end{figure}

\begin{figure}[t]
\centering
\includegraphics[width=0.36\textwidth,trim=67 500 100 30]{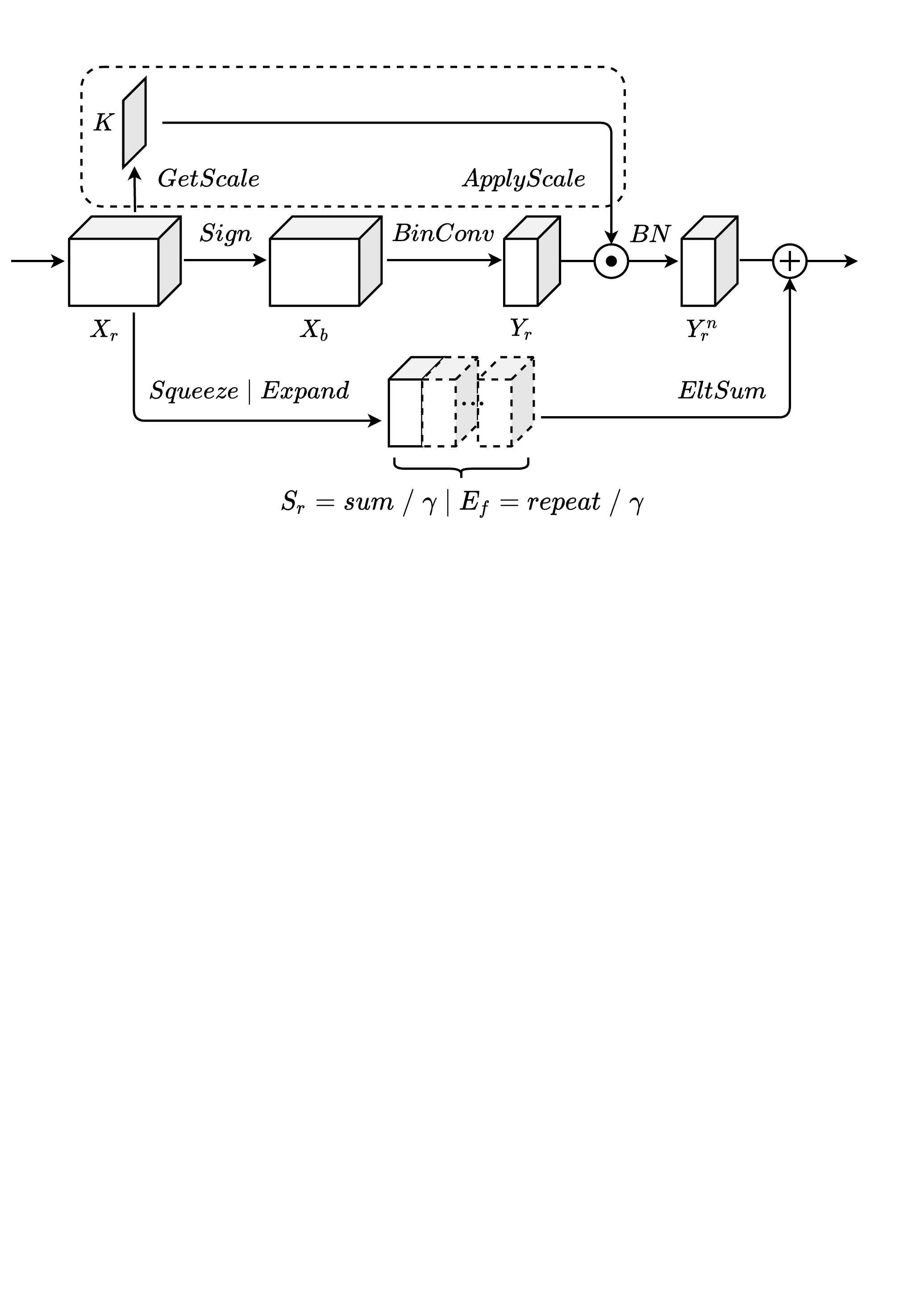}
\caption{Diagram of the Elastic-Link module. $\oplus$ denotes element-wise summation. The process of applying the scaling factor to activations is depicted within the dashed box, indicating that it is omitted in some cases for better performance. $GetScale$ and $ApplyScale$ operations refer to XNOR-Net~\cite{rastegari2016xnor}. }
\label{fig:EL-module}
\end{figure}

However, BNNs suffer from significant accuracy degradation as a result of information loss at each binarized layer. To conduct binarization, a real-valued signal is passed through the $Sign$ activation, which eliminates the signal's amplitude and retains only its sign information. Because this process is irreversible, information loss increases with each layer of the BNN. Therefore, a central challenge to improving BNN accuracy is the reduction of this information loss.

One approach seeks to minimize the quantization error between the real-valued and binary forms of the weights and activations. ~\citet{rastegari2016xnor} utilizes scaling factors to reduce the euclidean distances between the two forms. More recently, ~\citet{liu2018bireal} employs a sophisticated fine-tuning strategy from a full-precision network with customized gradient approximation methods. ~\citet{liu2018bireal} additionally proposes a shortcut connection to forward the real-valued activation, drastically reducing the extent of information loss. However, these methods apply best to networks which consist primarily of $3\times3$ or $5\times5$ convolutions, such as AlexNet~\cite{krizhevsky2012alexnet}, VGGNet~\cite{simonyan2014vgg} and Basic-Block ResNet~\cite{he2016resnet}.

Performing binarization with the above methods on networks in which $1\times1$ convolutions play a crucial role - for example, GoogleNet~\cite{szegedy2015googlenet}, Bottleneck ResNet or efficient networks with separated convolutions~\cite{howard2017mobilenets} - causes substantially greater accuracy degradations. $1\times1$ convolutions fuse information across channels and are already used to reduce computational cost via dimensionality reduction. We hypothesize that the marginal information loss from binarization is the proverbial last straw. ~\cite{howard2017mobilenets} observed that the training or fine-tuning of binarized MobileNet fails to even converge, giving credence to this hypothesis.

In order to make binarization more widely applicable, we introduce an effective and universal module named \textit{``Elastic-Link''} (EL). In order to compensate for the loss incurred by binarization, we adaptively add the real-valued input features (i.e. the features before feeding into the binarization function) to the output features of the subsequent convolution to retain the original real-valued signal. ~\citet{liu2018bireal} demonstrated that adding extra shortcut connections, implemented by an element-wise summation, on the Basic-Block ResNet~\cite{he2016resnet} produces considerable improvement in accuracy. This can be viewed as a special case of our proposed EL in which the input and output have the same shape. To generalize this finding, we develop a method to enable feature addition even if the feature size is changed by the convolution. EL uses a \textit{Squeeze} or \textit{Expand} operation to align the feature sizes between the input and the output. Furthermore, we do not simply perform a direct summation after the \textit{Squeeze} or \textit{Expand} operation, but rather learn a scaling factor to balance the relative extents of preserving the real-valued information and convolutional transformation, unifying these mechanisms and fusing them in a learnable, light-weight manner. EL is applicable to any architecture without structural limitations. To better visualize the effects of information preservation, we illustrate the feature-maps of a full-precision model, a Bi-Real model and our model with EL separately in Fig.~\ref{fig:heatmap}. The feature maps of our EL model show clear object contours and the retention of important information for recognition, with less noise. By contrast, the foreground and background in the Bi-Real model are not as easily discriminated. We believe that the EL module benefits information flow in the binarized neural networks.

Moreover, as shown in Fig.~\ref{fig:EL-module}, the design of the EL module is simple and can be directly applied to existing modern architectures. To assess the effectiveness of EL, we conduct extensive experiments on the ImageNet dataset. We outperform the current state-of-the-art result with a top-1 accuracy of 68.9\%. We also contribute comprehensive ablation studies and discussions to further understanding of the intrinsic characteristics of BNNs.
\section{Related Work}

\noindent\textbf{Quantized Weights with Real-Valued Activations.} Restricting weights to be either +1 or -1 allows the time-consuming multiply-accumulate (MAC) operations to be replaced with simple addition operations. Recent studies~\cite{courbariaux2015bc, rastegari2016xnor} rely on this insight and employ the straight-through estimator (STE)~\cite{bengio2013ste} to tackle non-differentiability in the back-propagation of gradients during training. Courbariaux et al. proposed BinaryConnect~\cite{courbariaux2015bc} which drastically decreases computational complexity and storage requirements while achieving good results on the CIFAR-10 and SVHN datasets, but lacks experiments on large-scale datasets like ImageNet~\cite{russakovsky2015imagenet}. In Binarized Weight Networks (BWN)~\cite{rastegari2016xnor}, the full-precision weights are binarized during each forward and backward pass on the fly, updating only the full-precision weights. BWN achieves a notable accuracy increase, especially on large-scale classification tasks. Finally, TWN~\cite{li2016ternary} and TTQ~\cite{zhu2016ttq} use ternary instead of binary weights \{$-\alpha^-$, 0, $\alpha^+$\} to pass more information.

\noindent\textbf{Quantized Weights and Activations.} Another approach which has recently gained popularity restricts both the weights and the activations to \{-1, +1\}. This allows the convolutional operations to be completely replaced by the efficient \textit{xnor} and \textit{bitcount} operations~\cite{courbariaux2016bnn, rastegari2016xnor}, thus gaining extreme efficiency. XNOR-Net~\cite{rastegari2016xnor} is one of the most representative works of this approach and achieved remarkable accuracy with various networks. Bi-Real net~\cite{liu2018bireal} introduced additional shortcuts to retain the information of the real-valued activations, so as to alleviate information loss from binarization. The paper additionally introduced a custom gradient approximation method with a sophisticated fine-tuning strategy to further increase accuracy. Recently, ~\citet{bethge2019back} demonstrated that the said fine-tuning strategy and gradient approximation methods are not necessary for training BNNs - even without these, the authors achieved superior accuracy training from scratch with the simple straight-through estimator. Another method proposed to counteract the information degradation phenomenon is the linear combination of multiple binary weights to approximate full-precision weights, as introduced by ABC-net~\cite{lin2017abc}. Finally, TBN~\cite{wan2018tbn} takes ternary inputs\{-1, 0, +1\} and binary weights with scale factors \{$-\alpha$, $\alpha$\}, demonstrating the method on both image classification and object detection tasks.


\section{Methodology}

In this section, we first revisit the standard process for training BNNs, then subsequently introduce a novel module, ``Elastic-Link'' (EL), to reduce the information loss in BNNs. Lastly, we demonstrate the EL module on Bottleneck ResNet~\cite{he2016resnet} and MobileNet~\cite{howard2017mobilenets}.

\subsection{Gradient Approximation}
It is standard to use the $Sign$ function to binarize a CNN. Real values are converted to the binary set of \{-1, +1\} by the following equation:
$$ Sign(x)=\left\{
\begin{aligned}
&+1 &if \quad x \geq 0 \\
&-1 &otherwise\\
\end{aligned}
\right.
\eqno{(1)}
$$

where $x$ refers to a real-valued weight or input/activation. To facilitate training, binarization is typically executed on the fly and only the real-valued weights are updated by the gradients, as described in ~\cite{courbariaux2016bnn, rastegari2016xnor}. During inference, the real-valued weights are unused and binary weights are used as a drop-in replacement.

In the backward pass, since the $Sign$ function is non-differentiable everywhere, an approximation is used. In this work, we follow the conventional ``straight through estimator'' (STE)~\cite{bengio2013ste} unless otherwise stated. The approximated gradient in STE is formulated as:
$$ \frac{\partial Sign(x)}{\partial x} =\left\{
\begin{aligned}
&1 &if -1 \leq x \leq 1 \\
&0 &otherwise\\
\end{aligned}
\right.
\eqno{(2)}
$$

\subsection{Scaling Factor for Weights and Activations} \label{sec:scale}

As proposed in XNOR-Net~\cite{rastegari2016xnor}, a binary convolutional operation can be given as follows:
$$
\begin{aligned}
BinConv\left(\mathbf{A}, \mathbf{W}\right) \approx\left(Sign\left(\mathbf{A}\right) \otimes Sign\left(\mathbf{W}\right)\right) \odot \mathbf{K} \alpha
\end{aligned}
\eqno{(3)}
$$

where $\mathbf{A} \in \mathbb{R}^{c \times h \times w}$ is the real-valued input activation and $\mathbf{W} \in \mathbb{R}^{c \times k_h \times k_w}$ is the real-valued convolutional kernel. Here ($c$, $h$, $w$, $k_h$, $k_w$) refer to \textit{number of input channels}, \textit{input height}, \textit{input width}, \textit{kernel height}, and \textit{kernel width} respectively. $\otimes$ denotes the efficient XNOR-Bitcounting operation~\cite{rastegari2016xnor} that replaces the time-consuming arithmetic operations.

$\alpha$ is a scaling factor given by a vector L1-Normalization $\alpha = \frac{1}{n}\|\mathbf{W}\|_{\ell 1}$, which helps minimize the L2 error between the real-valued weights and the binary weights with scalar coefficient $\alpha$. $K$ is a two-dimensional scaling matrix for the input activation, whose shape corresponds to the convolutional output. It is given by setting each element with the same principle as $\alpha$. XNOR-Net concluded that $\alpha$ is more effective than $K$, which can even be entirely ignored for simplicity due to the relatively small improvement realized. Similarly, ~\citet{liu2018bireal} and ~\citet{lin2017abc} validated the effectiveness of $\alpha$ across various networks and datasets. Recently, ~\citet{bethge2019back} found that these scaling factors did not result in accuracy gains when BatchNorm~\cite{ioffe2015bn} is applied after each convolutional layer. In our experiments, we did observe the same experimental phenomenon with the Basic-Block ResNet, which is constructed with only $3\times3$ convolutions. However, we find that this principle holds only for $3\times3$ convolutions. When binarizing $1\times1$ convolutions, the scaling factor is still influential, which will be elaborated on later section.

\subsection{Elastic-Link}

Inspired by the shortcut connection mechanism, we use an element-wise summation operation to add real-valued input features to the output features generated by a binary convolution. In our Elastic-Link module, instead of an identity shortcut, we apply either a \textit{Squeeze} or an \textit{Expand} operation when a convolution alters the feature shape.
A diagram of our proposed Elastic-Link module is shown in Fig.~\ref{fig:EL-module}.
Formally, let $X_r$ denote the real-valued input feature where $\mathbf{X_r} \in \mathbb{R}^{H_i \times W_i \times C_i}$. We binarize $X_r$ through a \textit{Sign} activation function and obtain the binary $X_b$. Next, a binary convolution and standard BatchNorm are applied to obtain the convolutional output feature $Y_r^n \in \mathbb{R}^{H_o \times W_o \times C_o}$.

In order to compensate for the information loss, we add the real-valued input $X_r$ to the normalized convolutional output $Y_r^n$. If the input size is equal to the convolutional output size, an identity shortcut connection with element-wise summation is applied as proposed in Bi-Real net~\cite{liu2018bireal}. However, this condition is rarely true. In practice, a convolution operation usually changes the number of channels, and occasionally changes the height and width as well.
In the channel-reduction case, we design a \textit{Squeeze} operation in which the real-valued input $X_r$ is split into multiple groups along the channel axis without overlap. The number of channels for each group is $\lceil \frac{C_i}{C_o} \rceil$. We additionally zero-pad the features on input $X_r$ to ensure that $C_i$ can be exactly divided by $C_o$. Next, we sum these feature groups together to yield the squeezed features which will be of the same shape as $Y_r^n$. To reduce the effect of amplitude increase from the summation, and to offer a self-balancing tradeoff between information preservation and transformation, we divide the squeezed feature by a learnable scalar $\gamma$ initialized as the number of groups.
We take a similar approach for the channel expansion case. In an \textit{Expand} operation, the real-valued input feature is repeated several times and then concatenated to match the feature size of the convolutional output. The expanded feature is correspondingly divided by same learnable factor of $\gamma$.
Finally, the output of the \textit{Squeeze} or \textit{Expand} operation is added to the convolutional output feature $Y_r^n$, giving the overall output of the binarized convolution module. If spatial downsampling is required, a $2\times2$ max-pooling with stride 2 is applied before \textit{Squeeze} or \textit{Expand} to ensure spatial compatibility. The Elastic-Link module is formulated as follows:
$$
\begin{aligned}
EL\left(\mathbf{X_r}, \mathbf{W}, \gamma\right) = BN\left(BinConv\left(\mathbf{X_r}, \mathbf{W}\right)\right) \\ + SEI\left(\mathbf{X_r}, \gamma\right)
\end{aligned}
\eqno{(4)}
$$

Where $\mathbf{X_r}$ refers to the real-valued input activation and $\mathbf{W}$ refers to the convolutional weight. $BN$ and $Sign$ refer to the BatchNorm and Sign function respectively. $SEI$ refers to $Squeeze$, $Expand$ or $Identity$ operation, depending on the ratio of input and output channels. $\gamma$ is the aforementioned learnable parameter that balances information preservation and transformation.
We initialize $\gamma$ by the following equation and optimize it through back-propagation:
$$ \gamma =\left\{
\begin{aligned}
&\lceil \frac{C_i}{C_o} \rceil ,&C_i >= C_o \\
&\lceil \frac{C_o}{C_i} \rceil ,&C_i < C_o \\
\end{aligned}
\right.
\eqno{(5)}
$$

Where $C_i$ and $C_o$ refer to the number of channels for the input and output of a convolution respectively. $\lceil \; \rceil$ denote the ceiling or round-up operation. The max-pooling operation in the downsample case as well as the additional ReLU for efficient networks are omitted here for clarity.

\begin{figure}
\centering
\includegraphics[width=0.24\textwidth,trim=140 350 150 30]{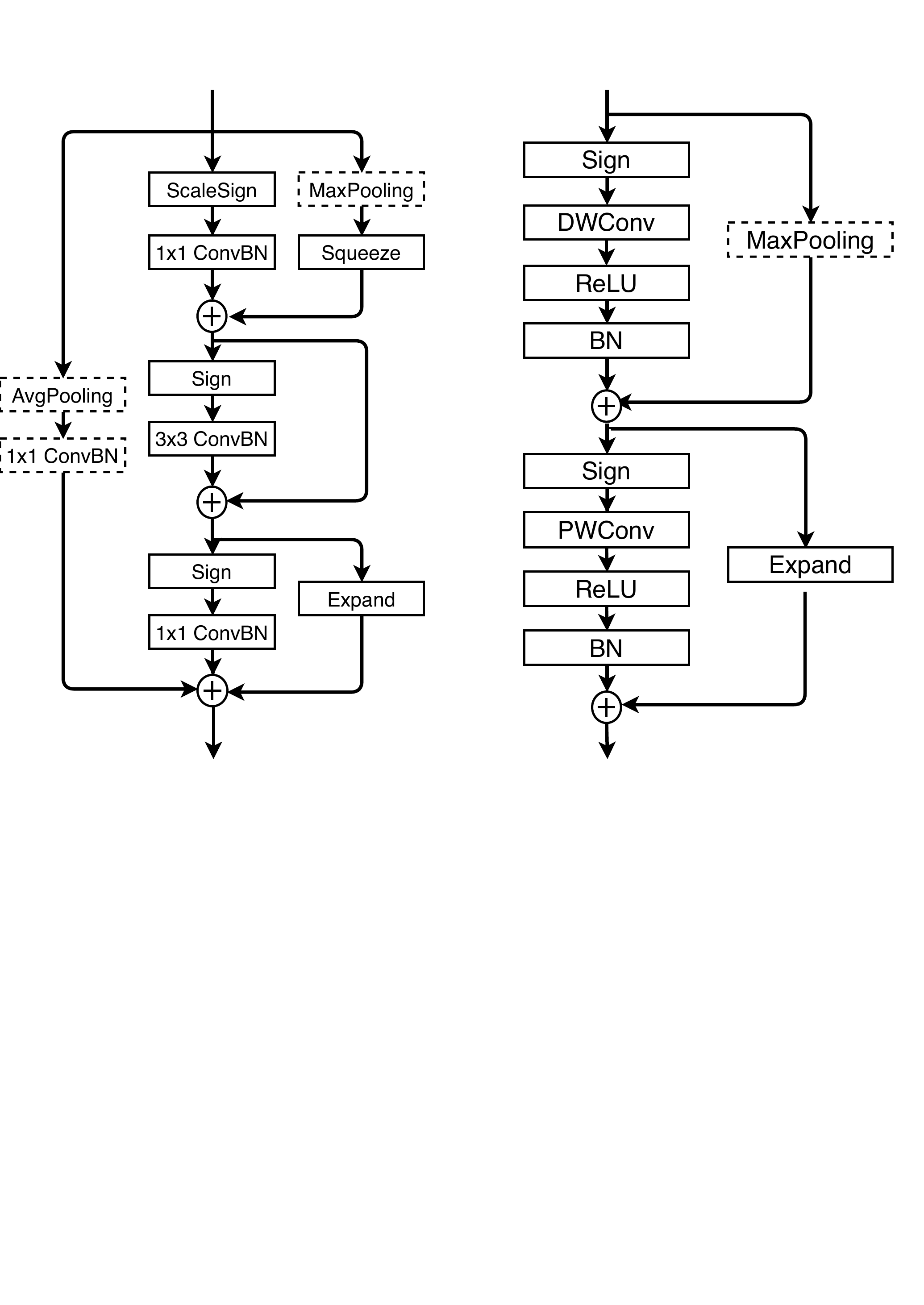}
\caption{Schema of EL-Bottleneck module (Left) and EL-MobileNet module (Right).}
\label{fig:EL-exampler}
\end{figure}

\noindent\textbf{Instantiations.} The Elastic-Link module easily plugs into many modern architectures. Taking Bottleneck ResNet as an example, we integrate the Elastic-Link module into ResNet26 which consists of 8 bottleneck blocks. All bottleneck blocks are replaced by EL-Bottlenecks, as depicted in Fig.~\ref{fig:EL-exampler}. The first convolution (of kernel size $7\times7$) and the classification layers remain full-precision to keep essential information at the input and output of the whole network. The downsampling shortcut in the first block of each stage, originally a $1\times1$ convolution of stride 2, is replaced by a $2\times2$ average pooling with stride 2 followed by a $1\times1$ convolution in full-precision. By integrating an EL module into the first $1\times1$ convolution of each \textit{Bottleneck} block, more full-precision information flows to the middle $3\times3$ convolution, which is essential for capturing features at larger receptive fields.

Next we apply the EL module to efficient networks composed of separable convolutions. To the best of our knowledge, efficient networks have so far been considered incompatible with binarization. MobileNet~\cite{howard2017mobilenets} is one of the most representative efficient architectures. By adding Elastic-Link to each pointwise convolution and depthwise convolution (see Fig.~\ref{fig:EL-exampler}), we are able to overcome the non-convergence problem typically encountered in training binarized MobileNet. We additionally find that keeping the ReLU activation achieves better performance. Similar to the ResNet case, we keep the first convolution, classifier and downsample components at full-precision.

\section{Computational Complexity}

For the Elastic-Link module to be considered practical, it must offer a good tradeoff between improved performance and additional computational burden. To illustrate the increased complexity associated with the Elastic-Link module, we compare Bi-ResNet50 with EL-ResNet50. The additional computational cost incurred by the EL module originates from the $\gamma$ scaling after the \textit{Squeeze}, \textit{Expand} or \textit{Identity} operation as well as the element-wise summation of each $1\times1$ convolution, because the \textit{Squeeze} and \textit{Expand} can be implemented with address mapping without any overhead. In total, EL-ResNet50 requires an extra $\sim$8M FLOPs over Bi-ResNet50's $\sim$300M FLOPs for a single forward pass with an input image of $224\times224$, corresponding to a 2.6\% increase. The number of FLOPs is computed as described in ~\cite{liu2018bireal}. For a practical comparison, we use the BMXNet library~\cite{haojin2017bmxnet} on an Intel Core i7-9700K CPU to measure the actual time taken. Bi-Real-ResNet50 takes on average 22.2 ms for a single forward pass (over 10 runs), compared to 22.9 ms for our proposed EL-ResNet50. We believe that this small additional cost is justified by the increase in model performance.

\section{Experiments}

In this section, we conduct extensive evaluations to demonstrate the effectiveness of Elastic-Link for binarized neural networks on the large scale image classification task ImageNet~\cite{russakovsky2015imagenet}. ImageNet is challenging and often used to validate the performance of proposed methods in BNNs. The dataset consists of about 1.28 million training images and 50 thousand validation images, annotated for 1000 classes.

\subsection{Training Details}
During training, we perform binarization on-the-fly in the forward pass and follow the STE gradient approximation strategy. Input images are resized such that the shorter edge is 256 pixels, then randomly cropped to $224\times224$ pixels, followed by a random horizontal flip. Mean channel subtraction is used to normalize the input.
All networks are trained from scratch using the Adam optimizer without weight decay normalization. The entire training process consists of 100 epochs with a mini-batch size of 256. The initial learning rate is set to 1e-3 and decreases after the 50th and 80th epochs by a factor of 10 each.

During inference, we center-crop patches of size $224\times224$ from each image on the validation set and report the top-1 accuracy for comparison.

\subsection{Effect of Scaling Activation} \label{sec:scaling_activation}
We have previously discussed the importance of the scaling weights and activations in binarized neural networks. Scaling is intended to mitigate the differences between the distributions of full-precision tensors and their binary counterparts. However, we obverse that scaling only the weights results in a minimal increase of accuracy when training from scratch, as previously observed in ~\cite{bethge2019back}. We expect that scaling the activations will greatly improve performance.

To investigate the effects of the scaling factor $K$ on activations, we conduct experiments on Bi-Real nets - binarized ResNet26 with an additional shortcut at the middle $3\times3$ convolution of each block - here named S-ResNet26 for convenience. ResNet26 is constructed from a set of homogenous \textit{Bottleneck} blocks which each comprises three convolutions: a first $1\times1$ channel-reduction convolution to reduce computation burden, a middle $3\times3$ convolution to capture spatial information, and a final channel-expansion $1\times1$ convolution to align the channel count with the input features for residual connection.

The experimental results are shown in Table~\ref{table:k}. Applying the scaling factor to only the first channel-reduction convolution of each bottleneck block results in a top-1 accuracy of 59.8\%, a significant increase over the original 58.7\% wherein no scaling factors are applied. When scaling is also applied to the other two convolutions of each block, a slight accuracy drop is observed. Based on this phenomenon, we speculate that the scaling factor for activations facilitates optimization of the channel-reduction convolution. Noting that improper application of the scaling factor can cause adverse effects, we apply scaling on only the first (channel-reduction) convolution of each block in our subsequent experiments on \textit{Bottleneck} ResNet.

\subsection{Integration with Basic-Block ResNet}
We next apply the EL module to the Basic-Block ResNet to validate its benefits. A Basic-Block is constructed by two successive isomorphic convolutions, and thus a feature passing through the block maintains its shape, except where downsampling is applied. In a Basic-Block, the EL module resolves to the identity shortcut connection as proposed in Bi-Real nets~\cite{liu2018bireal}. The binarized ResNet model constructed using Basic-Blocks with the EL module is almost the same as the Bi-Real net in terms of model architecture. However, the Bi-Real net utilizes a multi-stage fine-tuning strategy and the custom differentiable approximation ApproxSign, whereas the EL models are trained from scratch with simple STE. As can be observed in the results listed in table~\ref{tab:stoa}, both EL-ResNet18 and EL-ResNet34 are superior to their Bi-Real variants by a margin of 2.2\% and 1.0\% in top-1 accuracy respectively, demonstrating that a simple training schedule suffices to achieve a well-trained binarized model. 

\subsection{Integration with Bottleneck ResNet}
We next investigate the effectiveness of the proposed EL module on Bottleneck ResNet. Using the activation scaling strategy reported above, we obtain a strong baseline on S-ResNet26. Subsequently, we apply the EL module to each bottleneck block (see Fig.~\ref{fig:EL-exampler}) and conduct extensive ablation studies. The results are reported in Table~\ref{table:el_resnet26} and we can obverse the following phenomena:

\begin{itemize}

\item[-] From the result of model $EL_1$, $EL_2$ and $EL_3$, we can see that applying Elastic-Link on more convolutions within a network can monotonically bring benefits ($59.8\% \rightarrow 61.8\% \rightarrow 62.1\% \rightarrow 63.2\%$), demonstrating the effectiveness of the EL module.

\item[-] Comparing model $EL_3$ with $EL_4$, when the original residual shortcuts are removed throughout the network, there is a significant drop of top-1 accuracy from 63.2\% to 59.2\%. This proves that although EL provides more information via the additional connection for each convolution, residual connections are still essential to forward primitive uncompressed information.

\item[-] Comparing model $EL_6$ with $EL_3$, allowing $\gamma$ to be learnable further increases the top-1 accuracy from 63.2\% to 64.0\%.

\item[-] From the results of models $EL_5$ and $EL_6$, it can be observed that EL is compatible with the scaling factor strategy and achieves better accuracy when used with other techniques.

\end{itemize}

We term $EL_6$ in Table~\ref{table:el_resnet26} as EL-ResNet26, and this is constructed using EL-Bottleneck (see Fig.~\ref{fig:EL-exampler}). Including EL with learnable $\gamma$ in the convolutional layers results in dramatic gains, increasing top-1 accuracy by an absolute value of 4.2\%. 

\begin{table}[t]
\renewcommand\arraystretch{1.1}
\begin{center}
\setlength{\tabcolsep}{5mm}{
\begin{tabular}{cccc}
\hline
$K_s$ & $K_i$ & $K_e$ & Top-1 \\
\hline
           &            &            & 58.7 \\
           &            & \checkmark & 58.5 \\
\checkmark &            & \checkmark & 59.3 \\
\checkmark & \checkmark &            & 59.8 \\
\checkmark & \checkmark & \checkmark & 59.4 \\
\checkmark &            &            & \textbf{59.8}\\
\hline
\end{tabular}}
\end{center}
\caption{Top-1 accuracy (\%) of S-ResNet26 variants on ImageNet. $K_s$, $K_i$ and $K_e$ mean applying activation scaling to the first convolution (channel-reduction), middle convolution (spatial) and last convolution (channel-expansion) of each bottleneck block respectively.}
\label{table:k}
\end{table}

\begin{table}[htb]
\renewcommand\arraystretch{1.1}
\begin{center}
\setlength{\tabcolsep}{2mm}{
\begin{tabular}{lccccccc}
\hline
 & $K_s$ & $EL_s$ & $EL_i$ & $EL_e$& $Id.$ & $\gamma_l$ & Top-1 \\
\hline
Baseline&\checkmark  &       &            &            & \checkmark &            & 59.8 \\
$EL_1$&\checkmark & \checkmark &            &            & \checkmark &            & 61.8 \\
$EL_2$&\checkmark & \checkmark &            & \checkmark & \checkmark &            & 62.1 \\
$EL_3$&\checkmark & \checkmark & \checkmark & \checkmark & \checkmark &           & 63.2 \\
$EL_4$&\checkmark & \checkmark & \checkmark & \checkmark &            &            & 59.2 \\
$EL_5$&           & \checkmark & \checkmark & \checkmark & \checkmark & \checkmark & 63.6 \\
$EL_6$&\checkmark & \checkmark & \checkmark & \checkmark & \checkmark & \checkmark & \textbf{64.0} \\
\hline
\end{tabular}}
\end{center}
\caption{Top-1 accuracy (\%) of binarized ResNet26 with different configurations of EL on the ImageNet validation set. $EL_s$, $EL_i$ and $EL_e$ denotes applying Elastic-Link to the first convolution (channel-reduction), middle convolution (spatial) and the last convolution (channel-expansion) of each bottleneck block respectively. $Id.$ means keeping residual connections, and $\gamma_l$ means setting $\gamma$ to be learnable in the EL module. }
\label{table:el_resnet26}
\end{table}


\subsection{Results on Deeper and Efficient Networks}
In order to validate the generalizability of the EL module, we apply it to deeper networks, such as ResNet50~\cite{he2016resnet}, which are rarely reported in existing literature. The results in Table~\ref{tab:stoa} shows that EL-ResNet50 (65.6\% top-1 accuracy) is superior to the Bi-Real-ResNet50 (62.7\% top-1 accuracy) by a significant margin, proving that the EL module maintains strong performance even as the network grows deeper.

We next demonstrate the efficacy of the EL module on MobileNet. This is much more challenging as separable convolutions are weak at capturing spatial features. More concretely, the depthwise convolutions lack inter-channel information, while the pointwise convolutions which are expected to perform the inter-channel aggregation are particularly sensitive to information loss. As such, binarizing a MobileNet causes serious degradation in the network's ability to extract strong features. We add additional shortcut connection on depthwise convolutions and also perform the multi-stage fine-tuning strategy as discussed in Bi-Real net~\cite{liu2018bireal}. However, this binarized MobileNet still failed to converge in its training loss. Subsequently, we included the EL module into MobileNet (see Fig.~\ref{fig:EL-exampler}), and obtained the results shown in Table~\ref{tab:stoa}. The EL module not only enabled convergence, but also achieved an excellent performance of 56.4\% top-1 accuracy, a strong baseline for future work on binarized efficient networks.

\begin{table*}[t]
\renewcommand\arraystretch{1.1}
\centering
\setlength{\tabcolsep}{1.5mm}
\scalebox{0.96}{

\begin{tabular}{l|cc|c||cc}
\hline
   & \multicolumn{2}{c}{\underline{Bottleneck Block}} & \underline{Efficient Block} & \multicolumn{2}{c}{\underline{Basic Block}}  \\
   & RN26 & RN50 & MobileNet & RN18 & RN34 \\
\hline
Full-Precision  & 72.5 & 75.9 & 70.6 & 69.3 & 71.5 \\
XNOR~\cite{rastegari2016xnor} & 52.1 & 54.2 & \textit{Not Converge} & 51.2 & 53.2 \\
ABCNet~\cite{lin2017abc} & 45.2 & 52.9 & \textit{Not Converge} & 42.7 & 52.4 \\
TBN~\cite{wan2018tbn}       &  - & - & - &  55.6 & 58.2 \\
Bi-Real~\cite{liu2018bireal}  & 57.8 & 62.7 & \textit{Not Converge}  & 56.4 & 62.2 \\
BinaryE~\cite{bethge2019back} & 57.9 & 61.2 & \textit{Not Converge} & 56.7 & 59.5 \\
CI-Net~\cite{ziwen2019cinet} & - & - & - & 56.7 & 62.4 \\
XNOR++~\cite{adrian2019xnor++} & - & - & -  & 57.1 & -\\
GBCN~\cite{liu2019gbcn}    & - & - & - & 57.8 & - \\
MoBiNet~\cite{hai2020mobinet} & - & - & 54.4 & - & - \\
\textbf{EL (Ours)} & 64.0 & 65.6 & 56.4 & 60.1 & 63.2 \\
\hline
Real-to-Bin~\cite{martinez2020r2b} & 64.8 & 65.9 & 54.8 & 65.4 & 66.1 \\
\textbf{EL$^\dag$ (Ours)} & \textbf{67.1} & \textbf{68.9} & \textbf{61.2} & \textbf{65.7} & \textbf{66.5} \\
\hline

\hline
\end{tabular}
}
\caption{Comparison of top-1 accuracy (\%) on the ImageNet validation set. EL is specifically designed to improve $1\times1$ convolution, and its superiority compared to other works is correspondingly obvious across networks constructed with Bottleneck or Efficient Block. For reference, we also include networks built with Basic Block (which lack $1\times1$ convolutions). ``RN'' is short for ResNet. EL$^\dag$ refers to including Real-to-Bin~\cite{martinez2020r2b} in our proposed EL networks.}
\label{tab:stoa}
\end{table*}

\subsection{Comparison with the State-of-the-Art}
Finally, in order to demonstrate the superiority of our proposed EL in BNNs, we compare our results with other work on binary weights and activations. The main results on ImageNet are listed in Table~\ref{tab:stoa}. We demonstrate that binarized networks integrating with our EL module obtain considerable gains and also achieve the best performance against previous methods. At the same time, our approach integrates well with the Real-to-Binary~\cite{martinez2020r2b} approach, which is important for practical applications. Remarkably, we also integrated EL with the state-of-the-art result ReActNet-C~\cite{liu2020react} with top-1 accuracy of 71.4\% on the Reduction block by replacing the duplicate activation parts with our EL modules. We obtained a new state-of-the-art result with a top-1 accuracy of 71.9\%.

\section{Discussion} \label{Sec:discussion}

\subsection{Effect of Binary $1\times1$ Convolution}
An EL module propagates the real-valued input signal without a change of receptive field regardless of the change in channel depth. Similarly, a $1\times1$ convolution linearly fuses inter-channel information also without a change of receptive field. Our experiments have comprehensively demonstrated the effectiveness of the EL module as a solution to the information loss incurred by binary $1\times1$ convolutions. We next investigate whether the EL module is effective outside of $1\times1$ convolutions. We set up a comparison on S-ResNet26 by replacing the first convolution (channel-reduction) of each block with the following alternatives: 1) a full-precision convolution, 2) an EL module and 3) an EL module without the inner $1\times1$ convolution. The result in Table~\ref{tab:discuss1x1} shows that the model with the complete EL module reaches almost the accuracy of the model with the full-precision convolution. Removing the $1\times1$ convolution from the EL module still results in an absolute 1.2\% improvement in top-1 accuracy. We therefore conclude that between the two mechanisms of fusing inter-channel information by binary $1\times1$ convolution and forwarding real-valued information, the latter is more crucial to improve accuracy in BNNs.

%
%
%
%

\subsection{The Role of the $\gamma$ Factor}
\begin{figure}[htb]
\centering
\includegraphics[width=0.4\textwidth,trim=130 265 120 265]{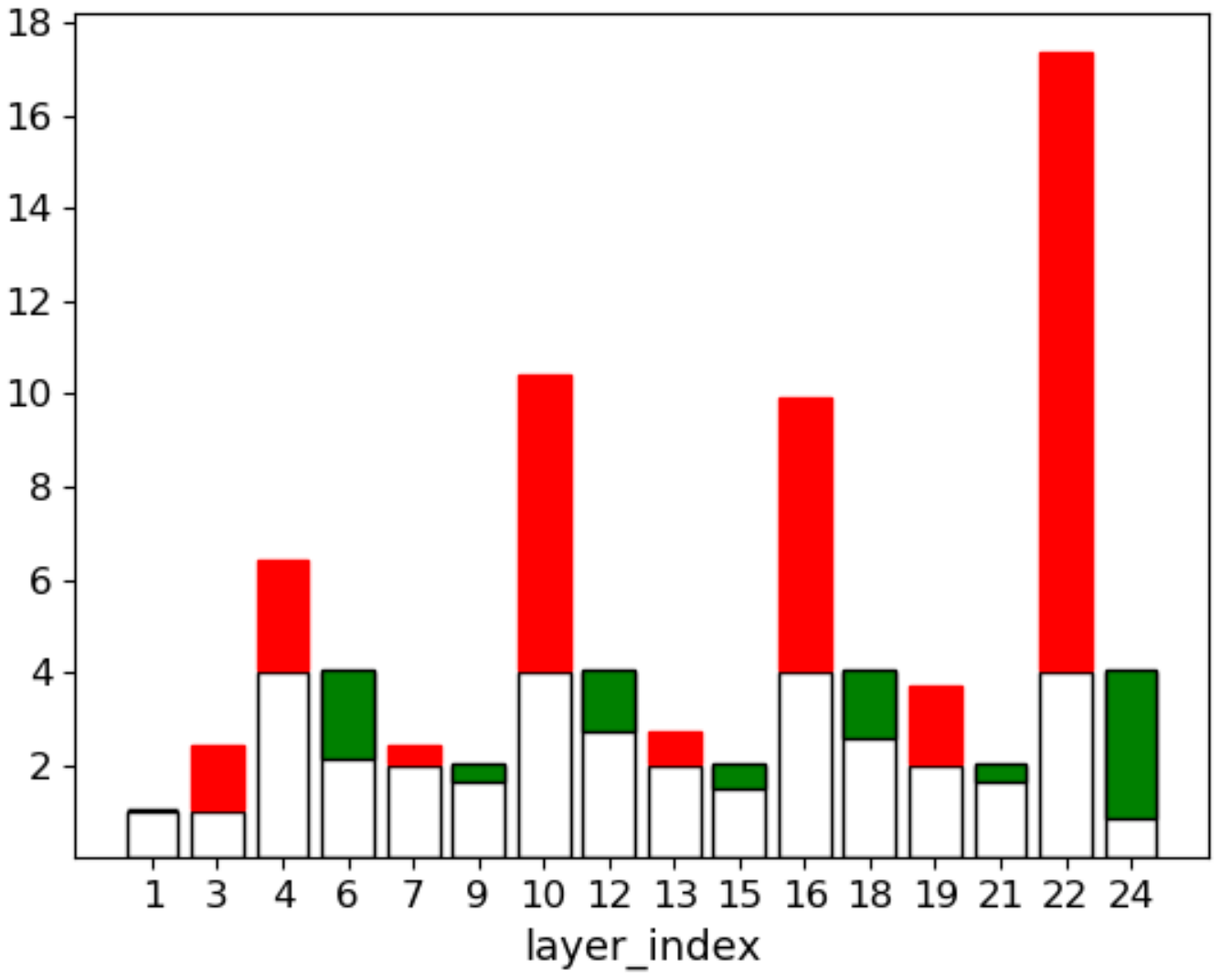}
\caption{Learnable factor $\gamma$ in all the EL module of EL-ResNet26 on ImageNet. The red histogram indicates an increase and the green histogram indicates a decrease compared to the initial value after training.}
\label{fig:gamma}
\end{figure}

The learnable factor $\gamma$ for the EL module controls the proportion of information fusion between the real-valued input features and the convolutional output features. In order to investigate the behavior of the $\gamma$ factor, we study the distribution of $\gamma$ from EL-ResNet26 with respect to the depth within the model. Fig.~\ref{fig:gamma} illustrates the value changes of all 16 $\gamma$ across 8 EL-Bottleneck blocks after training. Relative to their initial values, all the $\gamma$ in the EL module that were applied to the channel-reduction convolutions with kernel size $1\times1$ increased significantly. This indicates that less original information is retained as input into the subsequent spatial convolution. In contrast, the $\gamma$ values in the EL module that were applied to the channel-expansion convolutions slightly decreased, indicating that more information from previous spatial convolution is forwarded to the next block. This phenomenon is within expectations because if all the original information before a convolution is forwarded, the said convolution would be of relatively little utility as a feature transformer. It is also worth noting that all $\gamma$ applied to the channel-reduction convolutions in which downsampling occurred increased substantially during training (i.e. little original information is retained), as the convolutions in these shortcuts are full-precision, and thus are able to output more accurate information.

\begin{table}[tb]
\renewcommand\arraystretch{1.1}
\begin{center}
\setlength{\tabcolsep}{3.5mm}{
\begin{tabular}{cccc}
\hline
 FP $1\times1$ & Binary $1\times1$ & $EL^-$ & Top-1 \\
\hline
           &  \checkmark &            & 59.4 \\
\checkmark &             &            & 62.0 \\
           &             & \checkmark & 60.6 \\
           & \checkmark  & \checkmark & 61.8 \\
\hline
\end{tabular}}
\caption{Top-1 accuracy (\%) of various S-ResNet26 with different forms for the first convolutions of each block (the $1\times1$, channel-reduction ones). FP $1\times1$ refers to full-precision $1\times1$ convolution. Binary $1\times1$ refers to binarized $1\times1$ convolution. $EL^-$ refers to removing the $1\times1$ convolution from the original EL module, retaining only the \textit{Squeeze} operation.}
\label{tab:discuss1x1}
\end{center}
\end{table}

\section{Conclusion}

In this work, we proposed a novel Elastic-Link module for binarized neural networks. The Elastic-Link module introduces a connectivity mechanism to adaptively fuse real-valued input features and convolutional output features regardless of whether or not the feature-size is altered in the convolution. A learnable scaling factor enables an optimized tradeoff between information preservation and feature transformation, significantly reducing information degradation in the binarized form.
The Elastic-Link module can be easily embedded into any architecture. It greatly enhances the representational ability of BNNs, especially for networks in which $1\times1$ convolutions are indispensable, such as Bottleneck ResNet and MobileNet. Extensive experiments demonstrate that binarized networks with Elastic-Link achieve considerable performance gains with negligible computational overhead. Moreover, the module is compatible with other techniques that have been shown to improve accuracy, e.g. the application of a scaling factor to activations. Combining Elastic-Link with such techniques achieves a new state-of-the-art result. A key challenge remaining is the observed increase in the degree of information degradation with network depth. We plan to explore more effective approaches to counteract this phenomenon in future work.

\clearpage
\section{Acknowledgments}

The work is supported in part by NSFC Grants (62072449, 61972271), Macao FDCT Grant (0018/2019/AKP) and Guangdong-Hongkong-Macao Joint Research Grant (2020B1515130004).

{
\bibliography{egbib}
}

\end{document}